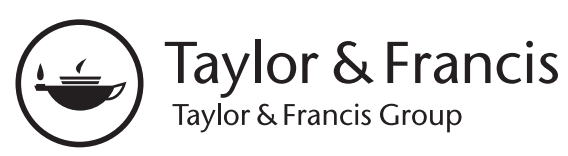

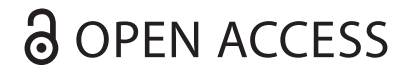
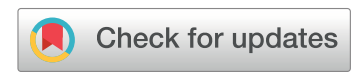

# Guiding Generative Storytelling with Knowledge Graphs

Zhijun Pan[a], Antonios Andronis[b], Eva Hayek[a,b], Oscar A. P. Wilkinson[b], Ilya Lasy[b], Annette Parry[b*], Guy Gadney[b*], Tim J. Smith[a#] and Mick Grierson[a#]

[a]Creative Computing Institute, University of the Arts London, London, UK; [b]Charismatic.ai, London, UK

**ABSTRACT**
Large language models (LLMs) have shown great potential in story generation, but challenges remain in maintaining long-form coherence and effective, user-friendly control. Retrieval-augmented generation (RAG) has proven effective in reducing hallucinations in text generation; while knowledge-graph (KG)-driven storytelling has been explored in prior work, this work focuses on KG-assisted long-form generation and an editable KG coupled with LLM generation in a two-stage user study. This work investigates how KGs can enhance LLM-based storytelling by improving narrative quality and enabling user-driven modifications. We propose a KG-assisted storytelling pipeline and evaluate it in a user study with 15 participants. Participants created prompts, generated stories, and edited KGs to shape their narratives. Quantitative and qualitative analysis finds improvements concentrated in action-oriented, structurally explicit narratives under our settings, but not for introspective stories. Participants reported a strong sense of control when editing the KG, describing the experience as engaging, interactive, and playful.

## 1. Introduction

Recent years have witnessed remarkable advancements in large language models (LLMs), spurring interest in automated content generation for applications such as interactive entertainment, creative writing aids, and educational tools (Akoury et al., 2023; Bryan-Kinns et al., 2024; Chakrabarty et al., 2024; Gallotta et al., 2024; Pan et al., 2025; Peng, Quaye, et al., 2024; Shaer et al., 2024; Sun, Li, et al., 2023; Taveekitworachai et al., 2023; Wang, Li, et al., 2024). Models like GPT-4 and Llama (Achiam et al., 2023; Touvron, Lavril, et al., 2023; Touvron, Martin, et al., 2023) have demonstrated the capability to produce coherent and engaging narratives in controlled settings (Beguš, 2024; Simon & Muise, 2022; Venkatraman et al., 2024). However, two persistent challenges remain. First, long-form story generation can lead to not only model hallucinations and inconsistencies in details but also broader issues in narrative quality, ranging from weak character development to disjointed plot progression, especially as the story grows in scope (Chen et al., 2023; Wang et al., 2023; Wang & Kreminski, 2024; Zhang et al., 2023). Second, direct and intuitive user control over the generative process is often limited, making it difficult for users to modify the story's content, pacing, or progression without intricate prompt engineering or laborious manual editing (Alabdulkarim et al., 2021; Alhussain & Azmi, 2022; Chakrabarty et al., 2024, 2025). In practice, prompt-only control can be difficult for non-experts and brittle across phrasings: users struggle to design and troubleshoot prompts (Zamfirescu-Pereira et al., 2023), and results can be sensitive to prompt selection, motivating multi-prompt evaluation protocols (Mizrahi et al., 2024). These concerns are frequently echoed in narrative-focused venues (e.g., ICIDS; Wordplay), where usability and evaluation stability are recurrent themes.

Existing solutions typically rely on extended textual prompts, iterative generation pipelines (e.g., chain-of-thought (CoT) reasoning; Wei et al., 2022), rule-based scripts, or in-text editing to guide

**CONTACT** Zhijun Pan a.pan@arts.ac.uk Creative Computing Institute, University of the Arts London, London, UK
*Both authors are the senior industry supervisors of this work.
#Both authors are the senior academic supervisors of this work.

LLM-based storytelling (Chakrabarty et al., 2024, 2025; Wang & Kreminski, 2024). While these methods can enhance coherence, they often become unstable in longer-form storytelling due to limited context windows of LLMs, which can lead to inconsistencies as the narrative progresses (Chen et al., 2023). Additionally, in terms of control, these approaches may be unintuitive for users to interact with, particularly nonprofessional writers who are unfamiliar with complex prompt-tuning techniques. This motivates structured, user-facing alternatives that externalize story state and afford direct manipulation beyond prompt text.

Meanwhile, although knowledge graphs (KGs) have proven valuable in areas such as question answering, recommendation systems, and information integration (Chen et al., 2020; Hogan et al., 2022), their narrative use has been explored in several strands of work, including graph-backed world/plot construction and retrieval-augmented interactive fiction (Ammanabrolu et al., 2020; Barros et al., 2019; Battad et al., 2019; Chambers et al., 2024; Yoo & Cheong, 2024). By structuring information into interconnected entities and relationships, KGs can systematically track story elements, mitigating hallucination and reinforcing narrative coherence. Moreover, their ability to represent complex story elements in a structured yet compressed form enables clearer progression and richer world-building. Crucially for usability, an editable KG offers a direct interface for end users to modify narrative elements at the structural level, potentially making control more transparent than prompt-only workflows.

Research on retrieval-augmented generation (RAG) (Gao et al., 2025; Lewis et al., 2020) and more recently graph RAG (Edge et al., 2024; Peng, Zhu, et al., 2024) has shown that augmenting LLMs with structured data can improve factual accuracy, consistency, and overall textual quality by grounding outputs in external knowledge sources to reduce model hallucinations. While most implementations focus on short-form tasks such as summarization or question answering, the core principles – dynamic retrieval, context maintenance, and user-driven updates – are highly relevant to storytelling, where long-form coherence and quality content remain critical goals.

Prior literature has also explored both the potential and limitations of integrating structured knowledge like KGs into story generation (Guan et al., 2020; Wang et al., 2023), as well as graph-structured interactive fiction and goal-driven systems (Ammanabrolu et al., 2020; Chambers et al., 2024; Yoo & Cheong, 2024). However, fewer studies examine end-user editing of a live graph during long-form generation with a user study focusing on perceived control and narrative quality. Moreover, comparatively little is known about when such editing feels effective to users across different narrative emphases. These gaps motivate our investigation of editable KGs as a structured alternative to prompt-only control, evaluated with users and analyzed across narrative types.

Building on these ideas, our research integrates KGs into LLM-based story generation to pursue two key objectives: producing higher-quality stories (e.g., richer character arcs, better plot development, and fewer hallucinations) and offering user-friendly control for editing story elements. Specifically, we propose a pipeline in which the story engine and KG modules communicate bidirectionally, ensuring that newly generated scenes align with previously established details. Users can modify the graph to introduce radical changes or subtle tweaks, instantly reflect those edits in regenerated or subsequent scenes. We then evaluate this approach in a two-stage study: Stage 1 (reading, no editing) and Stage 2 (editing), measuring story quality and perceived control. In analysis, we report an observed divergence between action-oriented and introspective narratives and introduce a post hoc grouping to contextualize results. To rigorously evaluate this approach, we formulated the following research questions:

*RQ1 (effectiveness of knowledge graph)*: How effective is the knowledge graph in enhancing the quality of generated long-form stories?

*RQ2 (control over story generation)*: Does the knowledge graph provide editors with a meaningful sense of control and agency during the story generation process?

We conducted a user study with 15 participants from a range of backgrounds, inviting them to create and edit stories across different genres and scenarios. We make three main contributions in this article:

1. We introduce a KG-based pipeline that enhances LLM-driven story generation, emphasizing longer-form narratives and user control.

2. We demonstrate, through user experiments, that editing KGs may support writer agency and creative freedom. This pattern is most evident in action-oriented story contexts.
3. We conduct both qualitative and quantitative evaluations of narrative quality and user satisfaction, revealing the strengths and limitations of our approach across different genres, and storytelling styles, and evaluation criteria.

## 2. Related work

### 2.1. Knowledge graph and retrieval augmented generation

Knowledge graphs organize data to aid machine understanding by connecting related information through defined entities and relationships. These graphs are widely used to improve search engine functions, enable personalized recommendation systems, and integrate information from various data sources (Chen et al., 2020; Hogan et al., 2022). Typically, creating a KG involves extracting relevant information from multiple sources and reasoning about this information to form meaningful connections. In narrative contexts, their use has extended beyond data integration to various forms of narrative support, including world and plot construction, interactive fiction tooling, and retrieval-augmented story systems (Ammanabrolu et al., 2020; Barros et al., 2019; Battad et al., 2019; Chambers et al., 2024; Yoo & Cheong, 2024).

The adoption of LLMs such as GPT-4 (Achiam et al., 2023) and Llama (Touvron, Lavril, et al., 2023; Touvron, Martin, et al., 2023) has greatly enhanced the development and utility of KGs. These models further automate aspects of KG construction by generating hypotheses, identifying new connections, and refining structures, thus enriching KGs with detailed, interconnected data that reflect complex real-world relationships (Kommineni et al., 2024; Meyer et al., 2023; Wang, Lipka, et al., 2024). Additionally, LLMs simplify the interaction with KGs by converting complex natural language queries into precise database queries, making these tools more accessible to non-experts and enhancing their practical applications (Zhu et al., 2024).

Within storytelling specifically, prior systems have leveraged structured graphs to scaffold plot/world construction and interactive play. For example, Barros et al. (2019) transform open data into graph-backed murder-mystery games; Battad et al. (2019) use multi-modal storytelling over cultural heritage collections; Ammanabrolu et al. (2020) generate interactive fiction worlds with graph-structured representations; Chambers et al. (2024) integrate retrieval-augmented state representations for interactive fiction; and Yoo and Cheong (2024) employ LLM-constructed graphs for goal-driven storytelling. Our contribution builds on this line by integrating an editable KG into an LLM pipeline and evaluating its end-user impact through a two-stage study that examines perceived agency and narrative quality in long-form generation.

In text generation, the functionality of KGs is expanded through RAG. This method allows LLMs to dynamically retrieve and integrate relevant information from knowledge bases during the text creation process, thereby improving the relevance, accuracy, and contextual depth of the text (Gao et al., 2025; Lewis et al., 2020).

Graph RAG is a more recent form of RAG that improves the retrieval process by using KGs to provide structured and contextually relevant information. It connects related entities from the graph to deliver more accurate and reliable outputs, addressing issues like hallucinations in LLMs by grounding their responses in external knowledge (Edge et al., 2024). This approach is particularly useful for tasks that require complex reasoning, such as summarization and question answering, where the system retrieves relevant parts of the graph to improve the quality of generated text. By using graph-based methods, Graph RAG creates more coherent and detailed content, improving both the factual accuracy and contextual relevance of the output (He et al., 2024; Peng, Zhu, et al., 2024). Microsoft has further automated this process by using LLMs, such as ChatGPT, to assist with KG extraction, entity generation, and retrieval inference, streamlining the pipeline (Edge et al., 2024).

Where prior KG-storytelling systems have largely emphasized system-side generation or designer-facing tooling, our study focuses on end-user editing of a live graph during long-form generation and measures its perceived controllability and effects on story quality.

### 2.2. Generative storytelling using large language models

Large language models are increasingly explored in storytelling applications such as scriptwriting, video games, and interactive media. These models generate dynamic narratives in real time, enabling player-driven experiences and co-creative storytelling systems (Akoury et al., 2023; Peng, Quaye, et al., 2024; Simon & Muise, 2022; Sun, Li, et al., 2023). Studies show that LLM-generated narratives can outperform human-curated stories in progressiveness and engagement under controlled conditions (Beguš, 2024; Zhao et al., 2023).

Not limited to textual modality, multimodal interactive storytelling combines text with visuals, offering richer narrative experiences. Some work has explored frameworks that generate visual stories that evolve alongside text, expanding possibilities in interactive media (Bensaid et al., 2021). LLMs have also been used in artistic installations to create personalized, evolving narratives based on user interaction (Sun, Tang, et al., 2023).

Recent research highlights that player perceptions of LLM-generated dialogue are influenced by factors such as coherence, immersion, and emotional engagement (Akoury et al., 2023). Due to token limitations of current LLM outputs, methods like CoT were introduced to generate longer outputs iteratively (Wei et al., 2022). To improve story quality, some approaches guide LLM-based generation to ensure logical flow and diversity in narratives (Wang & Kreminski, 2024).

Despite these advancements, LLM-driven storytelling faces several limitations. Inconsistency and hallucination in narrative flow are common issues, as models struggle to maintain logical coherence over long-form text [14, 17, 20] and alignment with intended user goals or stylistic expectations (Chakrabarty et al., 2025). Additionally, some studies suggest that LLMs may rely too heavily on learned patterns, producing stories that lack creativity, emotional depth, and playfulness (Beguš, 2024; Simon & Muise, 2022; Wang & Kreminski, 2024). Another issue is the generation of harmful or inappropriate content, particularly when LLMs are not properly guided or filtered (Taveekitworachai et al., 2023).

Beyond coherence and safety, prompt-only control remains challenging for non-experts and is sensitive to prompt phrasing variability. Empirical HCI work shows that many users struggle to design effective prompts and to troubleshoot failures (Zamfirescu-Pereira et al., 2023), while NLP evaluations caution that conclusions often depend on prompt selection and require multi-prompt protocols for robustness (Mizrahi et al., 2024). Related communities such as ICIDS and the Wordplay workshop have highlighted these usability and evaluation concerns in narrative applications. Motivated by this, we investigate editable KGs as a structured, user-facing alternative to purely prompt-based control.

### 2.3. Story evaluation criteria

The evaluation of narrative quality involves diverse subjective criteria, as proposed in established literary and narrative theories. To systematically understand and compare the quality of stories, particularly across various genres, researchers often utilize survey experiments guided by specific evaluation frameworks (Yang & Jin, 2024).

Evaluation methodologies focus on elements such as plot structure, character development, thematic clarity, and overall coherence. For instance, Freytag's model is utilized for analyzing plot effectiveness (Freytag, 1894), while character development is examined through practical writing guides (Bova, 1994). The clarity of a story's theme is assessed using literary standards set forth by Abrams (Abrams & Harpham, 2009), and narrative coherence is considered through Todorov's narrative theory (Todorov, 1971).

Other critical evaluation criteria include the emotional impact, language style, engagement, and intellectual stimulation of a story. Theories such as Holland's reader-response criticism (Holland, 1998) and Czikszentmihalyi's flow theory (Czikszentmihalyi, 1990) elucidate the emotional and cognitive engagement of the audience. Queneau's studies (Queneau, 2013) and Genette's narrative discourse framework (Genette, 1980) provide insights into the use of language and the intellectual depth of narratives. Cultural and social relevance are evaluated through Hall's encoding/decoding model, assessing how well stories reflect or resonate with contemporary social issues (Hall, 2019).

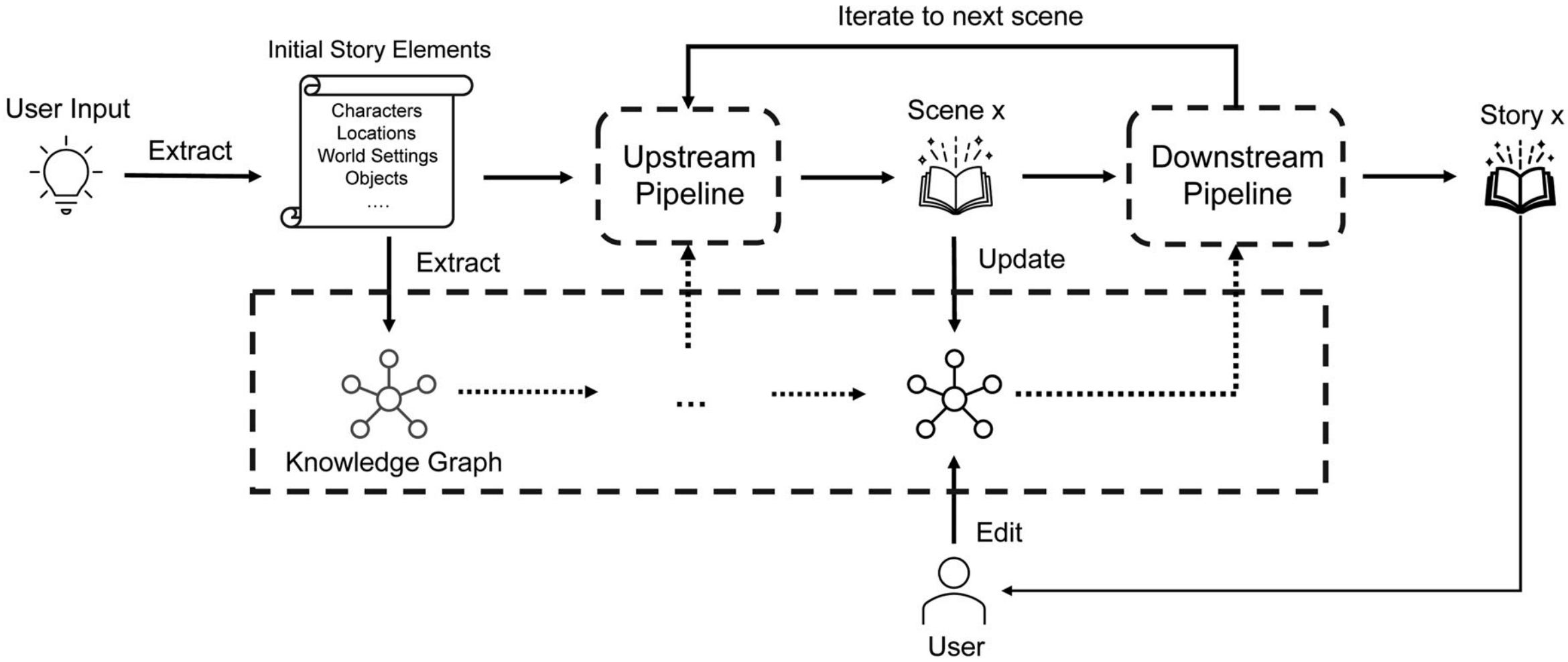

Figure 1. The story generation and knowledge graph pipeline (production version). *L1 state extraction* (user input → initial story elements; entities/relations extracted into the knowledge graph); *L2 retrieval/query* (the upstream pipeline consults the KG to condition drafting); *L3 scene generation* (scene *i* is produced); *L4 graph update* (facts from scene *i* update the KG and inform the downstream pipeline); *L5 optional edit/regeneration* (the editor modifies the KG; the system regenerates scene *i* or iterates to scene *i* + 1). The loop repeats per scene until the final story. This figure complements algorithm 1 and aligns with the stage 1/2 evaluation setup.

## 3. Methodology

### 3.1. Knowledge graph pipeline

Maintaining coherency and continuity while reducing hallucinations in extended narratives generated by LLMs presents significant technical challenges, particularly as stories grow in complexity and span multiple scenes. Common issues in long-form story generation include inconsistencies in geographic details, object properties, or character behavior. For example, a story might incorrectly describe a protagonist traveling by train from Europe to the USA, despite such routes being impossible within the established world settings. Similarly, character actions or object properties may contradict earlier parts of the narrative, resulting in a fragmented and disjointed reading experience.

To address these challenges, we have integrated a KG framework into our story generation pipeline (see Figure 1 and Algorithm 1), inspired by graph RAG (Edge et al., 2024). Graph RAG has proven effective in improving coherence and reducing hallucination across diverse text generation applications (Gao et al., 2025; Lewis et al., 2020). In our system, the KG acts as a central repository of narrative elements and relationships – such as characters, locations, objects, and events – while dynamically ensuring logical continuity and factual accuracy throughout the story.

Our KG pipeline is structured around three main stages: (1) *knowledge graph initialization*, (2) *context and scene generation*, and (3) *knowledge graph update*. These stages correspond to the symbolic procedures in Algorithm 1. Below, we describe how each stage functions, illustrating the step-by-step flow of data and operations.

*Knowledge graph representation.* The KG is represented using entries in the format "$A \rightarrow B \rightarrow R : D$", where $A$ and $B$ are nodes, and $R$ denotes the relationship connecting $A$ to $B$. $D$ provides a concise description associated with this relationship entry.

*Knowledge graph initialization.* At the outset, the pipeline takes user-provided inputs (e.g., $\mathcal{W}$ for world settings, story title, and a brief description) to initialize the node set $\mathcal{N}$ according to the user-defined types $\mathcal{T}$. Each type $\tau \in \mathcal{T}$ (e.g., *characters*, *locations*) is handled separately by the function

$$\mathcal{KG}_\tau = \text{ExtractKG}(\mathcal{N}, \tau, \mathcal{W}),$$

where $\mathcal{KG}_\tau \subseteq \mathcal{KG}$, which identifies and links the nodes relevant to that type. This step populates the knowledge graph $\mathcal{KG}$ with initial entities and relationships. Following this extraction, a call to $\text{CleanUp}(\mathcal{KG}, \mathcal{C})$ removes extraneous or duplicated entries from $\mathcal{KG}$, resulting in a streamlined, type-partitioned KG.

*Context and scene generation.* Once $\mathcal{KG}$ has been initialized, the pipeline iterates over $M$ scenes. At each iteration $i$, a scene-specific subset $\text{kg}_i \subseteq \mathcal{KG}_i$ is retrieved via

$$\text{kg}_i = \text{Query}(\mathcal{KG}_i, \mathcal{C}_{i-1}, \mathcal{S}_{i-1}).$$

If $i = 1$, the entire $\mathcal{KG}$ is used instead, since no prior scene $\mathcal{S}_{i-1}$ exists. Because the size of KGs grows as the story progresses, constraining the query to a relevant subset focuses the language model on essential relationships and elements, thereby minimizing token usage and reinforcing narrative coherence. The next scene $\mathcal{S}_i$ is then generated by

$$\mathcal{S}_i = \text{Generate}\left(\text{kg}_i, \mathcal{C}_{i-1}\right),$$

and the context $\mathcal{C}_i$ is updated to reflect the newly added narrative. This step-by-step approach ensures the story transitions smoothly and coherently between scenes, with each segment informed by the current KG subset, the previous scene, and the accumulated context.

*Knowledge graph update.* After each new scene is generated, the pipeline adjusts both the node set $\mathcal{N}$ and the global knowledge graph $\mathcal{KG}$. First,

$$\mathcal{N}_i = \text{UpdateNodes}(\mathcal{N}_{i-1}, \mathcal{S}_i)$$

adds or refines node attributes based on the latest scene's content. Then, each category of the knowledge graph $\mathcal{KG}_\tau$ is extended via

$$\mathcal{KG}_{\tau,i} = \mathcal{KG}_{\tau,i-1} \ \cup \ \text{UpdateKG}(\mathcal{KG}_{i-1}, \tau, \mathcal{S}_i),$$

introducing new relationships or augmenting existing ones with scene-specific details. Finally,

$$\mathcal{KG}_i = \text{CleanUp}(\mathcal{KG}_i, \mathcal{C}_i, \mathcal{S}_i)$$

removes redundant or no-longer-relevant information, keeping the KG concise and tightly aligned with the evolving story.

*Edit mode.* In addition to the standard update cycle, our system supports an optional *edit mode*, allowing users to revise $\mathcal{KG}$ after generating a scene. During this stage, users may open and modify the KG manually, then decide either to *regenerate* the same scene with the updated $\mathcal{KG}$ or to proceed to the next scene.

Like the story generation pipeline, our KG pipeline is designed to divide tasks into modular components, leveraging the strengths of LLMs for doing specific subtasks one at a time. This step-by-step approach enhances both the accuracy and usability of the generated content by allowing each component to focus on a distinct operation – such as node extraction, graph construction, or scene generation. This also facilitates iterative refinement and potentially promotes better contextual understanding at each stage of the pipeline, while reducing hallucination by controlling number of input tokens. By progressively integrating narrative elements, the pipeline ensures that newly generated content aligns seamlessly with previously established details, maintaining narrative coherence and reducing hallucination. The design of individual components for specific tasks also opens the potential for employing a multi-agent mixture-of-experts system, where each agent is fine-tuned to excel at its assigned subtask. For instance, dedicated agents could specialize in extracting specific narrative types, constructing relationships within the KG, or generating scenes with detailed contextual awareness.

### 3.2. Experiment setup

In production, we configure the KG pipeline as an external component interfacing with the story generation pipeline. For this experiment, we have further decoupled the KG from the story generation pipeline to allow isolated testing of its components and facilitate ablation studies (as in Algorithm 1). The system accepts user inputs like story title, genre, protagonists, and a brief description before generating the first scene. This setup involves implementing a downscaled version of both the upstream and downstream story generation pipelines that focus only on necessary minimized components and prompts for scene generation, respectively. This approach reduces complexity (minimizes prompt bias

**Algorithm 1** Knowledge Graph Pipeline for Story Generation

1: **Input:** $\mathcal{W}$: world settings, $\mathcal{T}$: user-defined types, $M$: number of scenes
2: **Variables:** Nodes $\mathcal{N}$, Knowledge Graph $\mathcal{KG}$, Context Summary $\mathcal{C}$, Scenes $\mathcal{S}$
3: **Initialize:**
4: $\mathcal{S} \leftarrow []$ of length $M$
5: $\mathcal{C} \leftarrow \emptyset$
6: $\mathcal{N} \leftarrow \text{InitializeNodes}(\mathcal{W}, \mathcal{T})$
7: **for** $\tau \in \mathcal{T}$ **do**
8: $\quad \mathcal{KG}(\tau) \leftarrow \text{ExtractKG}(\mathcal{N}, \tau, \mathcal{W})$
9: **end for**
10: $\mathcal{KG} \leftarrow \text{CleanUp}(\mathcal{KG}, \mathcal{C})$
11: **for** $i = 1$ to $M$ **do**
12: $\quad \text{kg} \subseteq \mathcal{KG} \leftarrow \begin{cases} \text{Query}(\mathcal{KG}, \mathcal{C}, \mathcal{S}[i-1]), & \text{if } i \neq 1, \\ \mathcal{KG}, & \text{otherwise.} \end{cases}$
13: $\quad \mathcal{S}[i] \leftarrow \begin{cases} \text{Generate}(\text{kg}, \mathcal{C}), & \text{if } i = 1, \\ \text{Generate}(\text{kg}, \mathcal{C}, \mathcal{S}[i-1]), & \text{otherwise.} \end{cases}$
14: $\quad \mathcal{C} \leftarrow \text{Summarize}(\mathcal{C}, \mathcal{S}[i])$
15: $\quad \mathcal{N} \leftarrow \text{UpdateNodes}(\mathcal{N}, \mathcal{S}[i])$
16: $\quad$ **for** $\tau \in \mathcal{T}$ **do**
17: $\quad\quad \mathcal{KG}(\tau) \leftarrow \mathcal{KG}(\tau) \cup \text{UpdateKG}(\mathcal{KG}, \tau, \mathcal{S}[i])$
18: $\quad$ **end for**
19: $\quad$ **if** *editMode* **then**
20: $\quad\quad$ **while** *true* **do**
21: $\quad\quad\quad \mathcal{KG} \leftarrow \text{Edit}(\mathcal{KG})$
22: $\quad\quad\quad$ **if** *nextScene* **then**
23: $\quad\quad\quad\quad$ **break**
24: $\quad\quad\quad$ **else**
25: $\quad\quad\quad\quad \mathcal{S}[i] \leftarrow \text{Regenerate}(\mathcal{S}[i], \mathcal{KG})$
26: $\quad\quad\quad\quad \mathcal{C} \leftarrow \text{Summarize}(\mathcal{C}, \mathcal{S}[i])$
27: $\quad\quad\quad$ **end if**
28: $\quad\quad$ **end while**
29: $\quad$ **end if**
30: $\quad \mathcal{KG} \leftarrow \text{CleanUp}(\mathcal{KG}, \mathcal{C}, \mathcal{S}[i])$
31: **end for**

and is better for analysis) and costs, enhances real-time generation speed, and increases flexibility, allowing for quicker and low-cost generation (*RQ1*) and dynamic adaptations like user edits to the KG, directly addressing the research question on user control over narrative development (*RQ2*).

For the implementation of the LLMs within the story and KG pipelines, we utilized GPT-4 for its stability in output generation and GPT-4o for enhanced processing speed. However, for this experiment, we have opted to use the smaller, lower-weight Llama 3.1 8B instead of GPT models. The decision is partly due to the non-open-source nature of GPT models, which limits our understanding of their internal workings and makes controlled ablation studies on the KG component more challenging. Llama 3.1, being smaller in size, allows for a more focused and transparent analysis of how the KG influences narrative development. This adjustment is crucial for assessing specific interactions between the narrative elements and the KG without the opaque processing layers typical of GPT models. To minimize model throughput and ensure consistency, we used basic prompts for story generation and KG extraction. Similarly, we used simplified prompts to describe the KG format for generation and usage purposes.

The system, including the story generation pipeline and KG pipeline, is hosted on a remote machine with an NVIDIA RTX A6000 GPU. We facilitate an online interactive platform for participants to engage in activities such as reading and comparing stories, editing the KG, and regenerating narratives based on their modifications (Table 1).

### 3.3. Participant workflow

*Participants.* We recruited $N = 15$ participants representing a diverse mix of academic, industry, and creative backgrounds. Roughly half were based in higher education institutions, including both technical disciplines (e.g., computing, engineering) and creative disciplines (e.g., design, communication), and covered roles from doctoral researchers to senior faculty. A number of participants were situated at the AI-arts boundary ("Creative AI"), spanning both academic and industry settings and including technical developers and arts-facing researchers. Several participants worked in creative industries such as writing, storyboarding, and production, while others were employed in AI and technology sectors, including engineers with a focus on creative applications of machine learning. In terms of expertise, some participants specialized in AI/ML and computational creativity, others in creative practice and production, and a smaller subset in broader engineering. Table 2 summarises this distribution across sector, role, and primary domain focus in anonymized form. Fourteen participants advanced to Stage 2, with one concluding after Stage 1 due to scheduling constraints.

**Table 1.** Nomenclature: Abbreviations, mathematical symbols, and study constants.

| *Abbreviations and labels* | |
|---|---|
| KG | Knowledge graph |
| LLM | Large language model |
| RAG | Retrieval-augmented generation |
| Graph RAG | Graph-based retrieval-augmented generation |
| CoT | Chain-of-thought (prompting) |
| K | *Kinetic narratives* (action/external emphasis; post hoc label) |
| I | *Introspective narratives* (interiority/reflective emphasis; post hoc label) |
| RQ | Research question |
| UI | User interface |
| *Symbols and notation* | |
| $M$ | Number of scenes per story (fixed to 5) |
| $\mathcal{W}$ | World settings/initial prompt context |
| $\mathcal{T}$ | Set of user-defined types (e.g., characters, locations) |
| $\tau$ | Type index in $\mathcal{T}$ |
| $\mathcal{N}$ | Node set (entities) |
| $\mathcal{KG}$ | Global knowledge graph; $\mathcal{KG}(\tau)$ denotes the type-partition |
| $\mathrm{kg}_i$ | Scene-$i$ subgraph retrieved for generation |
| $\mathcal{S}[i]$ (or $\mathcal{S}_i$) | Scene $i$ text |
| $\mathcal{C}$ | Rolling context summary |
| $P$ | Set of all participants ($\lvert P \rvert = 15$) |
| $P_A, P_B$ | Participants with valid ratings under conditions $A$ and $B$ |
| $P_{A,B}$ | Paired subset $P_A \cap P_B$ |
| $G$ | Subgroup (e.g., $K$ or $I$) |
| $P_{A,G}, P_{B,G}$ | Subgroup-specific subsets |
| $P_{A,B,G}$ | Paired subset within subgroup |
| $k$ | Number of Stage 1 criteria ($k = 8$) |
| $r_{p,i}(A)$ | Rating by participant $p$ for criterion $i$ under condition $A$ |
| $R_{\mathrm{agg}}(p, A)$ | Participant-level aggregated criterion score; see Equation (2) |
| $r_{p,\mathrm{hol}}(A)$ | Holistic (overall) rating under condition $A$ ; see Equation (4) |
| $d_p$ | Paired difference $R_{\mathrm{agg}}(p, A) - R_{\mathrm{agg}}(p, B)$ |
| $W^+, W^-, W$ | Wilcoxon's signed-rank positive/negative sums and test statistic |
| *Study constants and counts* | |
| $N$ | Total participants ($N = 15$) |
| Stage 1 ratings | $n = 30$ story versions rated (15 prompts $\times$ 2 conditions) |
| Post hoc groups | $N_K = 8$ (kinetic), $N_I = 7$ (introspective) |
| Scenes per story | $M = 5$ |
| Likert scale | 1 (Not Good) to 5 (Excellent) |
| Significance threshold | $p < 0.05$ (two-sided) |
| Model (experiment) | Llama 3.1 8B |
| Hardware | NVIDIA RTX A6000 GPU |

*Overview.* Each participant engages in the experiment on a one-to-one basis, with each session lasting approximately 60 min. The experiment consists of two primary stages: Stage 1 (*Editor Mode Off*), lasting around 20 min, and Stage 2 (*Editor Mode On*), lasting 40 min or longer if the participant wishes to extend their interaction. Stage 2 is independent of Stage 1. Throughout the experiment, participants input their own story prompts, generate stories using our system, interact with the KG, and evaluate the generated narratives.

The experiment is conducted on an interactive online platform that synchronizes the generated story output and the extracted KG. To ensure consistency across sessions, all participants follow the same structured workflow, with the only variable being the amount of time they choose to spend in Stage 2. By allowing participants to input their own story ideas, we introduce diversity into the generated narratives, mitigating biases in the model across different story genres and deepening our findings. Surveys for Stage 1 and Stage 2 were administered separately; before each survey, facilitators read a short script to standardize interpretation of the items.

*Briefing.* Prior to beginning the experiment, participants receive an introduction explaining the purpose and objectives of the study, along with a brief overview of the system and its KG integration.

**Table 2.** Participant backgrounds ($N = 15$), nested by sector.

| Category | Count |
|---|---|
| *Sector: academia/higher education* ($n = 8$) | |
| Roles | |
| Doctoral researchers | 4 |
| Research staff/faculty | 3 |
| Technical/engineering staff | 1 |
| Primary domain focus | |
| AI/ML (general) | 2 |
| Creative AI | 2 |
| Creative practice and production | 2 |
| Engineering (non-AI) | 2 |
| *Sector: industry (creative and tech)* ($n = 7$) | |
| Roles | |
| Creative professionals | 4 |
| Technical/engineering staff | 3 |
| Primary domain focus | |
| Creative AI | 3 |
| Creative practice and production | 4 |
| Total participants | 15 |

Within each sector, role and domain-focus counts are mutually exclusive and sum to the sector total; sector totals sum to $N = 15$.

They are explicitly informed that the system is separately designed solely for experimental evaluation and does not reflect the quality of a production-ready version. Facilitators read brief scripts before each survey and key task transition to ensure consistent understanding. Participants are then guided through the experimental procedure, including the expected duration and the two main stages:

- *Stage 1 (Editor Mode Off)*: Evaluates the effectiveness of KGs in enhancing AI-generated storytelling (*RQ1*). The participant will comparatively read and evaluate two stories generated using *identical* story prompts, where one story will involve a KG. Stage 1 will receive 2 responses per participant.
- *Stage 2 (Editor Mode On)*: Investigates whether allowing participants to edit the KG provides them with greater control over the generative process (*RQ2*). Independent from Stage 1, participants can use different story prompts to create their stories, edit KGs on each scene, choose to regenerate or proceed further to consequent scenes. Stage 2 will receive 1 response per participant.

Each generated story consists of five scenes, ensuring that the narratives are long enough to assess KG effectiveness in long-form storytelling while remaining manageable for participants to read within the allotted time. After a brief Q&A, participants access the interactive platform and the surveys.

*Stage 1: Editor Mode Off.* In Stage 1, participants provide their own story ideas through the interactive platform, including the story title, genre, protagonist details, and a brief description. To ensure clarity and safety, they are given an example story prompt and reminded to avoid generating violent, sexual, or otherwise inappropriate content, aligning with ethical guidelines and the built-in safety guard of the vanilla Llama model. Participants are also encouraged to provide concise yet sufficiently detailed descriptions to support effective story generation.

Two versions of the story are generated in this stage: one using a KG and one without. To minimize bias, participants are informed that the ordering of these versions is random. The story prompts remain *identical* for both versions to ensure controlled conditions and isolate the effect of the KG. After reading each story, participants complete a survey evaluating its quality. This stage is designed to address *RQ1*, assessing whether participants, under blinded conditions, perceive stories generated with KG assistance as superior or inferior, based on both quantitative ratings and qualitative feedback.

*Stage 2: Editor Mode On.* In Stage 2, we enable *Editor Mode*, allowing participants to directly modify the KG to influence story generation. We allow participants to edit their story prompts for this stage, as Stage 1 and Stage 2 are independent. Participants interact with the system after each scene is generated. The corresponding KG is displayed alongside the scene, enabling participants to review and edit it. They can freely add, remove, modify, and reconnect entries within the KG, adjusting relationships and narrative elements as they see fit. They are also encouraged to be creative in adding and modifying elements to make radical changes.

**Table 3.** Stage 2 (Editor Mode) survey instrument.

| Criterion | Construct | Response format |
|---|---|---|
| Perceived control | Self-reported sense of control in Stage 2; reflects perceived agency in the editing workflow. | Five-point Likert |
| Preference for editing system | Whether participants would choose to use the editing interface in future sessions. | Yes/no |
| Open-ended feedback | Qualitative reflections on the editing experience (e.g., changes attempted, what worked or failed, ideas for improvement). | Open-ended |

The original questionnaire items were: (1) "how much more in control do you feel of the generated story when you can edit the knowledge graph?" and (2) "would you prefer using the editing system or not using it (true/false)?". Facilitators explained the questions before the survey.

After making their modifications, participants have the option to either regenerate the current scene or proceed to the next one, both of which incorporate the edited KG. They can regenerate scenes as many times as they wish, and they may extend their participation time if they choose. For this stage, we administered a separate short survey focused on the experience during editing. Stage 2 emphasized qualitative feedback, and quantitative responses are reported descriptively.

This stage is designed to address *RQ2*, motivated by the need to characterize participants' experience while editing, including perceived agency and interactivity, using a stage-specific (descriptive) instrument. Unlike prior research on direct text-based interventions in AI storytelling, our approach examines the role of structured, element-driven modifications in shaping narrative outcomes, which we hypothesize may offer an intuitive pathway for non-expert editors (Mizrahi et al., 2024; Zamfirescu-Pereira et al., 2023). As *RQ2* primarily investigates the effectiveness of KG editing in influencing story content rather than the usability of the graph-editing interface itself, we intentionally kept the interface minimal and did not conduct a full usability evaluation; quantitative items are reported descriptively (see Table 3).

### 3.4. Evaluation framework and survey design

Evaluating stories is inherently subjective and challenging to measure objectively due to diverse reader preferences and interpretations. To ensure that our evaluation framework is both academically rigorous and professionally sound, we collaborated with storytelling professionals and academics. By integrating insights from industry partners, professional writers, and academic researchers, we grounded our framework in established and verified methodologies backed up by previous literature. This ensures that the evaluation criteria and survey design are robust, validated, and relevant to both academic and practical applications while backed up by previous research.

We administered separate, stage-specific surveys for Stage 1 and Stage 2. For both stages, facilitators read brief scripts immediately before the survey to standardize interpretation. The original question wordings are provided in Tables 3 and 4.

For *RQ1*, the evaluation criteria include the following dimensions: *Theme, Setting, Structure, Plot, Pace, Consistency, Characters, Dialogue, and Holistic (Overall Quality)*. For each of these criteria, participants rate the story using a five-point Likert scale (1 = Not Good, 2 = So-So, 3 = Good, 4 = Very Good, and 5 = Excellent). To complement the quantitative data and observe unforeseen patterns, an optional open-ended question invites participants to provide detailed qualitative feedback for both editor mode on and off. The full list of Stage 1 criteria and constructs is summarized in Table 4.

The evaluation is based on multiple interconnected criteria that assess both structural and artistic elements. Theme examines how effectively the story conveys its central message, ensuring coherence across narrative elements and emotional engagement, as outlined in Abrams' literary theory (Abrams & Harpham, 2009). Setting evaluates the richness and consistency of the story's environment, influencing immersion and believability, drawing from Ryan's study on narrative environments (Ryan, 2015) and Puxan-Oliva's research on spatial storytelling (Puxan-Oliva, 2024). Structure assesses the organization and progression of the story, ensuring logical transitions between scenes and acts, based on Freytag's Pyramid (Freytag, 1894) and McKee's principles of screenwriting (McKee, 1997). Plot measures the coherence and engagement of the sequence of events, examining conflict, resolution, and twists, grounded in Todorov's theory of narrative coherence (Todorov, 1971) and Brooks' narrative design framework (Brooks, 1992). Pace ensures that events unfold at a balanced rhythm, neither rushing nor stalling the narrative flow, following Gingrich's insights on narrative movement (Gingrich, 2021). Consistency evaluates the logical alignment of story details, character actions, and world rules, preventing contradictions that could break immersion, as discussed by Chatman and Chatman (1980) and Rimmon-Kenan (2003). Characters are assessed for their believability, development, and emotional depth, with Egri's (Egri, 1972) and Bova's (Bova, 1994) frameworks guiding how well character motivations and arcs contribute to the narrative. Dialogue is evaluated for its naturalness, narrative function, and ability to reveal character dynamics, drawing from Bakhtin's dialogic theory (Bakhtin, 2010) and Englert's work on structuring written discourse (Englert & Mariage, 1991). The Holistic criterion provides an overall assessment of story quality, reflecting the participant's general impression beyond specific narrative aspects.

**Table 4.** Stage 1 survey instrument.

| Criteria | Survey question | Ref. |
|---|---|---|
| Theme | How effectively does the story explore its central theme or main idea? | Abrams and Harpham (2009) |
| Setting | How well-developed and immersive is the story's setting, including its world-building elements? | Puxan-Oliva (2024) and Ryan (2015) |
| Structure | How much do you agree that the story is clear and well-organized (e.g., acts, plotlines)? | Freytag (1894) and McKee (1997) |
| Plot | How engaging and coherent is the sequence of events in the story? | Brooks (1992) and Todorov (1971) |
| Pace | To what extent is the story's pacing well-balanced? | Gingrich (2021) |
| Consistency | How consistent are the details, events, and characters throughout the story? | Chatman and Chatman (1980) and Rimmon-Kenan (2003) |
| Characters | How well-developed and believable are the characters in the story? | Bova (1994) and Egri (1972) |
| Dialogue | How natural and purposeful is the dialogue in the story? | Bakhtin (2010) and Englert and Mariage (1991) |
| Holistic | How would you rate the overall quality of the story? | |
| Open-ended | Optional free-response question for qualitative feedback. | |

For Editor Mode (Stage 2), the instrument prioritized qualitative insights into the editing experience (open-ended reflections), complemented by two brief items capturing perceived control during editing and a binary preference for using the editor in future sessions; given the qualitative emphasis, quantitative outcomes are reported descriptively (see Table 3).

Open-ended questions invited participants to provide qualitative feedback on both stages of the experiment. In Stage 1 (Editor Mode Off), participants could describe perceived differences between the two story versions (with and without KGs), highlighting strengths and weaknesses and explaining their preferences. In Stage 2 (Editor Mode On), participants reflected on their experience of using KGs for story control, focusing on their experience during editing (e.g., coherence, engagement, and creativity). This qualitative feedback provided valuable insights into the user experience, helping to identify additional patterns and analyze how participants interacted with AI-assisted storytelling.

### 3.5. Data analysis methods

To thoroughly analyze our results, we divide participant responses into different groups based on experimental conditions (e.g., use of a KG vs. no KG and various story genres). Each group may have a different number of responses because not all participants completed every stage and different participants chose different story genres. To properly compute statistical significance (using the Wilcoxon signed-rank test) for a pair of conditions, we only compare ratings from participants who provided responses in *both* conditions.

#### 3.5.1. Notation and setup

Let $P$ be the set of all participants in the study. Suppose we want to compare two conditions, $A$ and $B$. Define $P_A \subseteq P$ and $P_B \subseteq P$ as the subsets of participants who provided valid responses under conditions $A$ and $B$, respectively. For a fair paired comparison, we only consider:

$$P_{A,B} = P_A \cap P_B,$$

i.e., those participants who have valid ratings for both $A$ and $B$. For some cases, we further restrict the analysis to a *subgroup* $G \subseteq P$ of participants who meet a particular condition (e.g., those who selected a specific story genre). In that case, for each condition $A$ or $B$, we define:

$$P_{A,G} = P_A \cap G \quad \text{and} \quad P_{B,G} = P_B \cap G,$$

the sets of participants in subgroup $G$ who also provided responses under conditions $A$ or $B$. For a paired test within that subgroup, we then consider:

$$P_{A,B,G} = P_{A,G} \cap P_{B,G} \quad = \quad (P_A \cap G) \cap (P_B \cap G).$$

Hence, $P_{A,B,G}$ represents the participants *within subgroup* $G$ who provided valid responses for both conditions $A$ and $B$.

Each participant $p \in P_{A,B}$ provides ratings for $k$ criteria (in our case, $k = 8$) plus one overall holistic criterion. Let $r_{p,i}(A)$ or $r_{p,i}(B)$ denote the rating from participant $p$ on the $i$th criterion (where $i = 1, 2, \ldots, k$) under condition $A$ or $B$. Let $r_{p,\text{hol}}(A)$ or $r_{p,\text{hol}}(B)$ denote the holistic (overall) rating from participant $p$ under condition $A$ or $B$.

#### 3.5.2. Descriptive statistics

For each condition $A$ and each criterion $i$, we compute the mean rating $\overline{r}_i(A)$ and standard deviation $\sigma_i(A)$. Given $n_A = |P_A|$ valid responses under condition $A$, let the ratings for criterion $i$ be $\{r_{p,i}(A) : p \in P_A\}$. Then:

$$\overline{r}_i(A) = \frac{1}{n_A} \sum_{p \in P_A} r_{p,i}(A), \quad \sigma_i(A) = \sqrt{\frac{1}{n_A} \sum_{p \in P_A} \left(r_{p,i}(A) - \overline{r}_i(A)\right)^2}. \tag{1}$$

*Aggregated criterion rating.* We aggregate each participant's $k$ individual criteria into a single composite value. For participant $p$ in condition $A$, define:

$$R_{\text{agg}}(p, A) = \frac{1}{k} \sum_{i=1}^{k} r_{p,i}(A). \tag{2}$$

Hence, the group-level mean of the aggregated criterion rating in condition $A$ is:

$$\overline{R_{\text{agg}}}(A) = \frac{1}{n_A} \sum_{p \in P_A} R_{\text{agg}}(p, A). \tag{3}$$

*Holistic rating.* We treat the single holistic overall rating $r_{p,\text{hol}}(A)$ analogously by taking its mean and standard deviation across $p \in P_A$:

$$\overline{r}_{\text{hol}}(A) = \frac{1}{n_A} \sum_{p \in P_A} r_{p,\text{hol}}(A). \tag{4}$$

When sufficient data are unavailable (e.g., only a few participants fall under a particular condition or subcondition), we may report only these descriptive statistics (i.e., mean and standard deviation). Such cases arise when formal inferential tests lack statistical power or when participants in the subgroup are too few.

#### 3.5.3. Significance testing: Wilcoxon's signed-rank test

To examine whether a condition (e.g., "with knowledge graph" vs. "without knowledge graph") has a statistically significant impact on participant ratings, we use the Wilcoxon signed-rank test (Woolson, 2005) on paired data. Specifically, for conditions $A$ and $B$, we look only at participants in $P_{A,B} = P_A \cap P_B$.

*Setup for the test.* For each participant $p \in P_{A,B}$, define the difference in aggregated ratings:

$$d_p \;=\; R_{\text{agg}}(p, A) \;-\; R_{\text{agg}}(p, B).$$

Similarly, $d_p$ for each of the individual criteria or for the holistic rating can be defined.

*Wilcoxon's signed-rank calculation.* To compare two conditions, we first compute the paired differences $d_p$ for each participant in $P_{A,B}$. Any cases where $d_p = 0$ are excluded. Next, we rank the absolute differences $|d_p|$ in ascending order. The sum of ranks corresponding to positive differences ($d_p > 0$) is denoted as $W^+$, while the sum of ranks for negative differences ($d_p < 0$) is $W^-$. The Wilcoxon test statistic is given by:

$$W = \min(W^+, W^-).$$

Under the null hypothesis that there is no difference between conditions $A$ and $B$, $W$ follows a known distribution. The corresponding $p$ value is reported to determine statistical significance.

### 3.6. Post hoc kinetic versus introspective split

The distinction between kinetic and introspective narratives was introduced post hoc to account for consistent patterns observed in participants' prompts, edits, and feedback (see Section 4). In our operationalization, stories that foreground explicit goals, observable actions, and environmental change are contrasted with stories centered on inner states, subtext, and evolving relationships. This contrast aligns with evidence that LLMs are more reliable under explicit, checkable constraints (Gao et al., 2024; Jiang et al., 2024), and remain challenged by subtextual inference, theory-of-mind reasoning, and long-form faithfulness (Gómez-Rodríguez & Williams, 2023; Kim et al., 2024; Strachan et al., 2024; Subbiah et al., 2024). Grounded in narrative theory, we frame this as action/externality versus interiority/reflection (Chatman & Chatman, 1980; Zunshine, 2006). This grouping is exploratory and post hoc, added to explain observed differences rather than to test a preregistered hypothesis.

Both the human coder and an auxiliary LLM followed the same natural-language instructions. For each story, they considered the genre labels, the title, other participants' story prompts, and the story content to determine the predominant narrative emphasis. The abstracted decision rule and edge-case handling were expressed in natural language as follows:

**Definition.** Label the narrative kinetic when progression is primarily driven by explicit objectives, observable actions, spatial traversal, environmental change, or concrete task structure (e.g., heists, quests, exploration, puzzles, and combat). Label it introspective when progression is primarily driven by reflection, memory, inner conflict, subtext, relational negotiation, or affective/motivational change not tightly anchored to external tasks.

**Decision rule.** Judge predominant emphasis over all five scenes. If mixed, assign the label that best describes what moves the story forward and connects scene transitions.

**Edge cases.** If there is an explicit task structure but scene development hinges on inner thoughts, dilemmas, or relational subtext, prefer introspective. If emotions are rich but tightly yoked to stepwise task execution or spatial progression, prefer kinetic.

One human coder applied these instructions to all stories. To check that the instructions were interpretable and stable, we also ran an auxiliary classification pass with GPT-4o, selected for robust instruction following and long-context handling. The LLM's outputs served only as a non-binding sanity check; all analyses rely on the human labels. For clarity, we provide below an abridged template of the prompting setup:

```
messages = [{
        "role": "system",
        "content": """
                ROLE: Narrative analyst.
```

```
                RULES: < RULES >
                INSTRUCTIONS: Classify the five-scene story as
                kinetic or introspective using the rules above.
                Reply with JSON only:
                {
                    ”label”: “kinetic|introspective”,
                    ”keywords”: [“ < cue1 > “, “ < cue2 > “, “ < cue3 > “, .],
                    ”rationale”: “ < = 1 sentence”
                }”““
},
{
        ”role”: “user”,
        ”content”: “““
                GENRE LABELS: < labels >
                TITLE: < title >
                STORY DESCRIPTION: < text >
                PROTAGONISTS: < text >
                STORY BODY: < full text > “““
}]}
```

The model was required to output only a JSON object of the following form, as per this example:

```
{
    ”label”: “kinetic”,
    ”keywords”: [“quest”, “raid”, “combat”, “hacking”],
    ”rationale”: “Progression is driven by explicit objectives,
    team actions, and stepwise task execution across scenes.”
}
```

## 4. Experiment results and analysis

### 4.1. Overview

The results of our experiment are interesting (see Table 5). For *RQ1*, across all trial entries, the effectiveness of using KGs to assist story generation was insignificant. However, we observed that in certain story genres, particularly those rich in action-oriented or mechanically driven elements, the use of KGs substantially improved story ratings across the board, showing its statistically significant effectiveness. For *RQ2*, participants reported a strong sense of control while editing the KG and expressed clear interest in using it to steer generative storytelling. Under certain conditions, they exhibited high levels of playfulness, immersion, and creativity.

### 4.2. All stories ($N = 15$)

For *RQ1*, we received a total of $n = 30$ participant responses with 15 diversified story prompts under both conditions: story generation not assisted by KGs (*All*) and assisted by KGs (*All (KG)*). Overall, in terms of average rating and aggregated rating over all trials in the group, the KG version has a slightly

**Table 5.** Mean ratings with standard deviations in parentheses for different story conditions.

| Group | Theme | Setting | Struct. | Plot | Pace | Consist. | Char. | Dialogue | Holistic | Aggr. |
|---|---|---|---|---|---|---|---|---|---|---|
| All | 3.73 (1.03) | 3.40 (1.24) | 3.53 (0.92) | 3.80 (1.08) | 3.27 (1.16) | 3.73 (1.03) | 3.13 (1.13) | 3.33 (1.29) | 3.47 (0.99) | 3.49 (0.82) |
| All (KG) | 3.60 (1.12) | 3.53 (0.99) | 3.33 (1.11) | 3.80 (1.42) | 3.27 (1.22) | 3.53 (0.92) | 3.53 (1.13) | 3.47 (1.06) | 3.27 (1.16) | 3.51 (0.89) |
| K | 3.38 (1.19) | 3.00 (1.31) | 3.38 (1.06) | 3.50 (1.31) | 2.88 (1.36) | 3.50 (1.20) | 2.38 (0.74) | 3.00 (1.41) | 3.25 (1.16) | 3.13 (0.86) |
| K (KG) | 3.63 (1.41) | 3.63 (1.06) | 3.75 (1.28) | 4.13 (1.36) | 3.75 (1.39) | 3.63 (0.92) | 3.75 (0.89) | 3.63 (1.19) | 3.63 (1.30) | 3.73 (1.03) |
| I | 4.14 (0.69) | 3.86 (1.07) | 3.71 (0.76) | 4.14 (0.69) | 3.71 (0.76) | 4.00 (0.82) | 4.00 (0.82) | 3.71 (1.11) | 3.71 (0.76) | 3.91 (0.58) |
| I (KG) | 3.57 (0.79) | 3.43 (0.98) | 2.86 (0.69) | 3.43 (1.51) | 2.71 (0.76) | 3.43 (0.98) | 3.29 (1.38) | 3.29 (0.95) | 2.86 (0.90) | 3.25 (0.68) |

Aggr: the aggregated score across the eight evaluation criteria; All: includes all story trials ($N = 15$) without genre classification; K: kinetic narratives ($N = 8$); I: introspective narratives ($N = 7$); KG: stories generated with knowledge graph assistance.

higher aggregated rating compared to the version without (3.51 > 3.49), as well as higher ratings in specific criteria such as Setting (3.53 > 3.40), Dialogue (3.47 > 3.33), and a significantly higher rating in Characters (3.53 > 3.13). However, for other criteria and the Holistic rating, the KG version does not perform better. Across criteria, holistic, and aggregated ratings, the effectiveness of the KG is insignificant, with $p > 0.05$.

In qualitative feedback, 11 out of 15 participants explicitly stated that KGs added more details to the story; however, four of them (from the N group) expressed that this made the stories too literal, bland, and less vivid. As we are using a smaller 8B model, these findings might be due to the limitations of the model's capabilities and our decision to minimize pipeline components to eliminate model prompt bias, etc. However, from our observations, the performance of KGs varies significantly depending on the story types and genres.

### 4.3. Narrative grouping

During analysis of Stage 1 ratings and accompanying free-text comments, we observed a consistent divergence: stories foregrounding explicit objectives and observable actions tended to receive higher ratings under KG assistance, whereas stories emphasizing inner states and reflection tended to receive lower ratings. A similar pattern appeared in Stage 2: edits that instantiated concrete, externally verifiable changes (e.g., adding obstacles, altering locations, and adjusting goals) were more reliably incorporated than edits aimed at subtext or evolving relationships.

To interpret these outcomes, we introduced a post hoc grouping of stories into two categories and then re-summarized results by group. The grouping procedure, operational definitions, and edge-case rules are detailed in Section 3.6. The labels are used here solely to contextualize the observed differences; they were not part of a preregistered analysis plan.

### 4.4. Kinetic narratives ($N = 8$)

For stories rich in action-oriented and mechanically driven elements, where information is largely external, KG-assisted versions (*K (KG)*) outperform those without (*K*). We classify participants who did such stories into this category with a population size of $N = 8$, which typically includes genres such as Heist, Action Adventure, Sci-fi, Exploration, Dungeon-Crawling, and Combat. Across all evaluation criteria, including holistic and aggregated ratings, stories generated with KGs consistently received higher ratings. The effect of using KGs in this category is statistically significant, with $p = 0.039$ ($< 0.05$).

Notably, for the Character criterion, KGs proved highly effective, yielding a statistically significant improvement ($p = 0.016, < 0.05$). Similarly, the Pace and Structure criteria also showed notable improvements, with significance levels of $p = 0.053$ and $p = 0.083$, respectively. This trend is further supported by qualitative feedback, where in four out of eight trials, participants explicitly mentioned that KGs enhanced character interactions, descriptive details, internal thoughts, development, and overall believability. These improvements translated into substantially higher ratings for Characters (3.75 > 2.38) and Dialogue (3.63 > 3.00).

Beyond richer details and better characterization, four participants also highlighted that the KG version generated more engaging and well-structured stories. They reported improvements in story development and progression, reflected in significantly higher ratings for Pace (3.75 > 2.88), Plot (4.13 > 3.50), and Setting (3.63 > 3.00). Additionally, three participants provided deeper insights into how KGs enhanced storytelling logic, resulting in stronger narrative mechanics, more coherent action-adventure sequences, and a more complete story structure. These factors contributed to rating improvements across multiple criteria. However, views on story consistency remained mixed, with only a slight improvement in Consistency (3.63 > 3.50). This may be attributed to the limitations of the smaller 8B model, which struggles with maintaining coherence in longer texts, though the KG still facilitated some improvement.

These findings highlight the potential of KGs in enhancing story generation, particularly in genres characterized by physical movement, dynamic interactions, challenges, and conflicts. Such genres, which rely heavily on spatial awareness, terrain descriptions, quest items, mission objectives, and dynamically changing environments, benefit from the structured tracking capabilities provided by KGs. Additionally, the descriptive nature of KG versions does not hinder narrative immersion in these story types; instead, it enhances engagement. When participants can edit the KG, the storytelling experience becomes even more immersive and interactive.

### 4.5. Introspective narratives ($N = 7$)

The limitations of using KGs become evident in introspective narratives (*I*), which are rich in emotion-driven and psychologically complex elements, where information is largely internal rather than external. This category includes genres such as psychological horror, thriller, romantic drama, and other character-driven narratives that focus on internal conflicts, nuanced emotions, and implicit motivations. The effectiveness of using KGs (*I (KG)*) in these genres is close to statistical significance with $p = 0.07$ ($> 0.05$), but in a negative direction, indicating a decline in aggregated ratings ($3.25 < 3.91$). This suggests that, rather than aiding the generative process, KGs may inadvertently hinder the creation of such story types.

Participants in this group reported a variety of issues across different trials, including a lack of character autonomy, bland storytelling, overly literal descriptions, and poor story pacing. Notably, four out of seven participants explicitly described the story progression as repetitive and structurally rigid, a pattern that is nearly statistically significant with $p = 0.068$. These findings can largely be attributed to the limitations of the smaller 8B model, which struggles with processing implicit, abstract, and ambiguous information – key components in introspective storytelling. Unlike action-oriented genres, where concrete objects and explicit relationships drive the narrative, introspective stories rely on psychological depth, subtle emotional cues, and evolving character arcs, which the KG framework fails to effectively capture.

Overall, these results suggest that KGs may not be well suited for introspective narratives when using smaller LLMs, as they impose an external logic onto stories that are inherently driven by internal, often unstructured thought processes.

### 4.6. Playing with the knowledge graph ($N = 14$)

A total of 14 participants advanced to Stage 2, where they freely edited their KGs for each scene and chose whether to regenerate the current scene or proceed to the next one. The editing experience was well received: 13/14 participants (92.9%) indicated they would prefer to use the editing system in similar tasks. Perceived control during editing was rated highly: 3 participants (21.4%) marked “Excellent”, 8 (57.1%) “Very Good”, 2 (14.3%) “Good”, and 1 (7.1%) “Not Good” (see Figure 2).

Among the participants, five out of eight who engaged with *Kinetic Narratives* demonstrated remarkable creativity and playfulness when editing their KGs. They tested a wide range of imaginative changes, including altering physical obstacles, tactical environments, and character dynamics in ways that effectively impacted the narrative’s course and complexity. One participant, for instance, approached the story in a DnD-like fashion by adding traps, disabling power in crucial areas, and encrypting doors, while another staged a confrontation with an evil scientist by breaking the protagonist’s weapon, altering the laboratory layout and scientist’s motivation, and summoning reinforcement to save them by the end. Three others deliberately introduced comedic or absurd twists, such as turning a battle into a birthday party and a sudden popcorn fight, making a character endlessly repeat their friend’s words, or introducing random story elements. According to their qualitative feedback, these users found it gratifying to see the system successfully integrate their edits in real time coherently with existing context, elevating both immersion and their sense of agency in shaping the unfolding action.

All five of these participants reported high enjoyment, with some sessions lasting over two hours. Four specifically highlighted that the system accurately reflected or even completely redirected the storyline based on their inputs, indicating a strong alignment between their creative intentions and the final narrative. Two participants suggested that additional guidance – such as tutorials or example modifications – could further enhance the editing experience. Overall, their feedback suggests that KG editing can serve as an accessible, playful tool for storytelling, easing the burden of traditional writing while maximizing creative freedom. This potential is especially valuable for nonprofessional writers and interactive storytelling contexts where real-time narrative experimentation is desired.

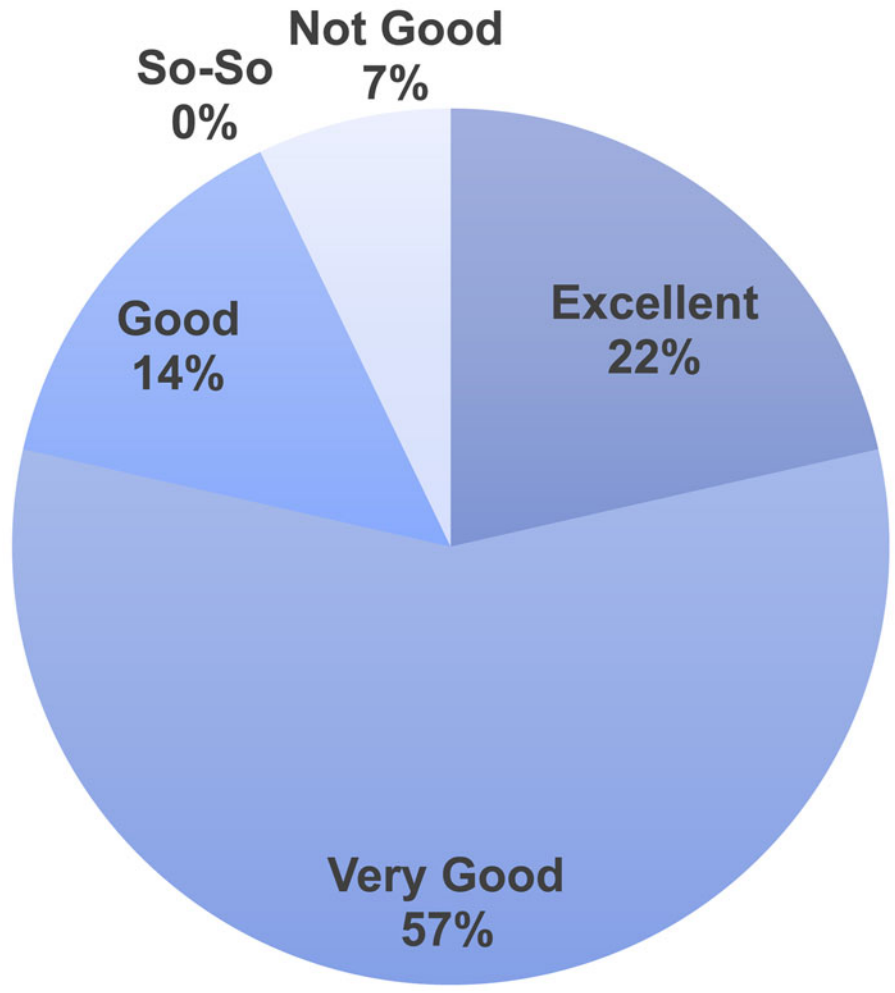

**Figure 2.** Distribution of participant ratings on their perceived level of control when editing the knowledge graph in Stage 2.

Nevertheless, the system also exhibited limitations in aligning with user intentions. As reported by seven out of 14 participants, some intended changes were either not reflected accurately or disrupted the narrative flow. We suspect that these issues were partially attributed to the small model's limited capacity to interpret implicit or nuanced user inputs – challenges that were pronounced in Introspective Narratives, which demand a deeper understanding of psychological subtleties. In comparison, such discrepancies were less evident in Kinetic Narratives, where the external and literal nature of elements allowed for straightforward modifications.

Closer inspection shows that these difficulties arose most clearly in participants' attempts to introduce more introspective or implied changes. By way of example, we observed cases where edits intended to convey sustained ambiguity (such as unease or a sense of being watched) were rendered as explicit entities or declarative statements. Similarly, edits concerning interpersonal nuance (e.g., latent resentment or shifting trust) were often expressed as overt dialogue, losing the subtlety participants had intended. In other cases, perspective or memory manipulations (e.g., unreliability, competing accounts) did not propagate across scenes, appearing instead as isolated asides. Finally, inner conflict (e.g., competing motives) sometimes surfaced only as self-reflection without influencing subsequent events, stalling progression. While only illustrative, these examples highlight a broader pattern: introspective intents were more coherently realized when re-expressed as observable proxies – such as environmental cues, timing mismatches, or decision points – that the model could more readily integrate into unfolding action.

These observations suggest that while KG editing offers a playful and immersive experience, its effectiveness varies significantly with the narrative type and the underlying model's capabilities. Future experiments with larger models and more diverse participant groups are necessary to explore this feature's potential and refine its application across different narrative styles.

## 5. Discussion and limitations

Our findings indicate that KGs can significantly enhance story generation in action-oriented and mechanically driven scenarios (*Kinetic Narratives*), but their benefits diminish or even reverse in introspective, emotionally focused genres (*Introspective Narratives*). We did not anticipate such a stark divide, suggesting that the nature of the story's content – external versus internal – substantially affects how well a KG can support narrative coherence and depth. However, it is important to note that our overall sample size was relatively small, limiting the statistical power of our significance tests. Further studies with larger participant groups are needed to confirm and refine these observations. We note that the kinetic-introspective split was introduced post hoc to interpret observed differences; in future work, we plan to preregister these criteria and obtain multi-coder reliability.

*Nature of the information.* A potential explanation lies in the distinction between external, concrete elements and internal, abstract dimensions. Kinetic Narratives draw heavily on physical objects, spatial arrangements, and clear character objectives, all of which map well onto a graph structure. In contrast, introspective stories hinge on subtle emotions, psychological growth, and nuanced motivations that are difficult to represent in a purely externalized, node-edge format. Future work should consider new ways of modeling internal states – perhaps through sentiment tracking, implicit relationship mapping, or graph schemas designed to capture psychological shifts – so that KGs can better accommodate introspective and complex storytelling.

*Model size and prompt influence.* We deliberately used minimal prompts and a smaller (8B) model to isolate the effects of the KG. This approach highlighted specific strengths and weaknesses, but it also constrained the system's overall performance. Larger or more capable models might raise general narrative quality in all story types, potentially concealing the discrete impact of the KG. Another consideration is the influence of prompt design: more detailed or context-specific prompts could benefit stories with intense internal complexity. For instance, carefully crafted instructions about character emotions, hidden motives, and psychological arcs may help smaller models or KGs handle introspective stories more effectively.

*Survey instrument design.* Stage 2 primarily prioritized qualitative feedback, complemented by two brief items on perceived control during editing and editor preference. Although facilitators used short scripts to standardize interpretation, there is a potential risk that the original control item could be read as implicitly comparative. In larger-scale studies, where efficiency and clarity become critical, such risks may matter more. Future work will therefore consider refining the questionnaire wording (e.g., piloting alternative phrasings, adding attention checks, and adopting validated scales where appropriate).

*Participant creativity and target users.* Many participants in Kinetic Narratives were notably playful and creative, raising questions about how individual creativity shapes the perceived usefulness of KGs. In our experiment, such participants are showing their playfulness since the beginning when creating their story prompts, long before the editing stage. Two of these participants emphasized the value of training and inspiration to help users exploit the system's creative potential. For some, guided tools, tutorials, or preset scenarios could prove more understanding on the system, especially if they prefer structured support rather than open-ended experimentation. It also emerged that one participant rated the editor experience as "Not Good", which may reflect mismatched expectations or a desire for different functionalities. Our findings hint that highly creative users – such as gamers, dungeon masters, or amateur writers – might enjoy experimenting with radical plot directions, while others may expect more straightforward or guided authoring. This suggests KG editing could find natural adoption in interactive story-based games, collaborative writing tools, or creative-writing workshops where some level of structural guidance is prized.

*Population and trial length.* Our sample size was small and drawn from a limited pool, leaving open questions about the generalizability of our conclusions. Additionally, the time constraints of our sessions, along with the small model's limited capacity, occasionally led to mismatched or missing user edits. This concern was more pronounced in introspective stories, where implied or abstract user changes proved challenging for the model to process. Future research should explore a broader, more varied participant base and systematically investigate the impact of editorial time, participant creativity, and model size on outcomes. In addition, the five-scene story length may have also influenced results. While sufficient to observe short-term effects, it may not fully reflect the long-term benefits of KGs in maintaining consistency over extended narratives. The lack of a noticeable improvement in consistency ratings suggests that the advantage of structured tracking might become clearer in longer story arcs.

*Implications and future directions.* Despite these constraints, KG editing shows promise as a fun, accessible, and powerful way to shape AI-generated stories. Participants valued the sense of control it afforded, particularly in fast-paced, action-driven contexts. Larger-scale studies and more advanced models could further test whether KGs maintain or increase their benefits when dealing with more ambiguous or introspective material. Incorporating more sophisticated editing interfaces and robust prompt strategies may help users convey abstract or nuanced story elements. Ultimately, refining these techniques, whether in interactive entertainment, creative writing support, or other domains, will reveal the full potential of KGs to balance user agency with narrative coherence, even in stories that delve into complex psychological terrain. A useful direction for future work is to compare knowledge-graph editing with alternative, non-KG interactive workflows (e.g., outline/beat editors, direct text edits, prompt templates or sliders) to assess effectiveness and usability across different narrative types. We also see value in refining the Stage 2 instrument to further minimize potential ambiguities, making it more robust for larger samples while continuing to preserve the richness of qualitative feedback.

## 6. Conclusions

In this article, we presented a KG-based pipeline to enhance long-form story generation using LLMs, addressing both narrative quality and user control. Findings from our user study, supported by both quantitative and qualitative data, indicate that KGs can significantly enhance story quality in kinetic narratives – stories characterized by action-oriented and mechanically driven elements. For such narratives, KGs enabled editors to modify story elements in an intuitive and playful manner, and participants described the experience as immersive and enjoyable.

However, introspective narratives, which rely heavily on psychological depth, nuanced character development, and implicit motivations, benefited less from the KG approach, highlighting the limitations of our method. This study represents an initial step toward integrating structured knowledge representations into generative storytelling. Future research will explore more capable models, refine the storytelling pipeline, and develop an advanced KG user interface. Additionally, we plan to conduct larger-scale studies with a more diverse participant population to further assess and extend the applicability of this approach. Future work will include systematic comparisons between knowledge-graph-based control and alternative, non-KG editing workflows to quantify benefits and tradeoffs.

## Acknowledgements

*Declaration of generative AI and AI-assisted technologies in the writing process*. During the preparation of this work, the authors used ChatGPT (OpenAI) for language refinement and proofreading. After using this tool, the authors reviewed and edited the content as needed and take full responsibility for the content of the published article.

## Author contributions

CRediT: **Zhijun Pan**: Conceptualization, Data curation, Formal analysis, Investigation, Methodology, Project administration, Software, Supervision, Validation, Visualization, Writing – original draft, Writing – review & editing; **Antonios Andronis**: Conceptualization, Methodology, Software, Validation, Writing – review & editing; **Eva Hayek**: Conceptualization, Data curation, Software, Validation, Writing – review & editing; **Oscar A. P. Wilkinson**: Software, Validation, Visualization, Writing – review & editing; **Ilya Lasy**: Conceptualization, Methodology, Validation, Writing – review & editing; **Annette Parry**: Investigation, Project administration, Resources, Supervision, Validation, Writing – review & editing; **Guy Gadney**: Conceptualization, Investigation, Project administration, Resources, Supervision, Validation, Writing – review & editing; **Tim J. Smith**: Formal analysis, Project administration, Resources, Supervision, Validation, Writing – review & editing; **Mick Grierson**: Conceptualization, Data curation, Formal analysis, Funding acquisition, Investigation, Methodology, Project administration, Resources, Supervision, Validation, Writing – review & editing.

## Ethical approval

This study was reviewed and approved by the College Research Ethics Subcommittee (CRESC) of the Creative Computing Institute, University of the Arts London (Ref: CCIS251). The research involved minimal risk and included surveys and interviews with professional screenwriters. All procedures, content, and data management protocols were reviewed and found to meet ethical standards.

## Consent form

Informed consent was obtained from all participants prior to their involvement.

## Disclosure statement

No potential conflict of interest was reported by the author(s).

## Funding

This work was supported by Innovate UK [Grant number 10104117].

## ORCID

Zhijun Pan http://orcid.org/0000-0003-4010-3648
Tim J. Smith http://orcid.org/0000-0002-2808-9401
Mick Grierson http://orcid.org/0000-0002-6981-5414

## References

Abrams, M. H., & Harpham, G. G. (2009). *A glossary of literary terms*. Wadsworth Cengage Learning.

Achiam, J., Adler, S., Agarwal, S., Ahmad, L., Akkaya, I., Aleman, F. L., Almeida, D., Altenschmidt, J., Altman, S., Anadkat, S., Avila, R., Babuschkin, I., Balaji, S., Balcom, V., Baltescu, P., Bao, H., Bavarian, M., Belgum, J., Bello, I., … Jain, S. (2023). *Gpt-4 technical report*. arXiv preprint arXiv:2303.08774

Akoury, N., Yang, Q., & Iyyer, M. (2023). A framework for exploring player perceptions of LLM-generated dialogue in commercial video games. In H. Bouamor, J. Pino, & K. Bali (Eds.), *Findings of the Association for Computational Linguistics: EMNLP 2023* (pp. 2295–2311). Association for Computational Linguistics. https://doi.org/10.18653/v1/2023.findings-emnlp.151

Alabdulkarim, A., Li, S., & Peng, X. (2021). *Automatic story generation: Challenges and attempts*. arXiv preprint arXiv:2102.12634

Alhussain, A. I., & Azmi, A. M. (2022). Automatic story generation: A survey of approaches. *ACM Computing Surveys*, *54*(5), 1–38. https://doi.org/10.1145/3453156

Ammanabrolu, P., Cheung, W., Tu, D., Broniec, W., & Riedl, M. (2020). Bringing stories alive: Generating interactive fiction worlds. In L. Lelis & D. Thue (Eds.), *Proceedings of the AAAI Conference on Artificial Intelligence and Interactive Digital Entertainment* (pp. 3–9). AAAI Press. https://doi.org/10.1609/aiide.v16i1.7400

Bakhtin, M. M. (2010). *The dialogic imagination: Four essays*. University of Texas Press.

Barros, G. A. B., Green, M. C., Liapis, A., & Togelius, J. (2019). Who killed Albert Einstein? From open data to murder mystery games. *IEEE Transactions on Games*, *11*(1), 79–89. https://doi.org/10.1109/TG.2018.2806190

Battad, Z., White, A., & Si, M. (2019). Facilitating information exploration of archival library materials through multi-modal storytelling. In R. E. Cardona-Rivera, A. Sullivan, & R. Michael Young (Eds.), *International Conference on Interactive Digital Storytelling* (pp. 120–127). Springer.

Beguš, N. (2024). Experimental narratives: A comparison of human crowdsourced storytelling and AI storytelling. *Humanities and Social Sciences Communications*, *11*(1), 1–22. https://doi.org/10.1057/s41599-024-03868-8

Bensaid, E., Martino, M., Hoover, B., & Strobelt, H. (2021). *FairyTailor: A multimodal generative framework for storytelling*. arXiv preprint arXiv:2108.04324

Bova, B. (1994). *The craft of writing science fiction that sells*. Writer's Digest Books.

Brooks, P. (1992). *Reading for the plot: Design and intention in narrative*. Harvard University Press.

Bryan-Kinns, N., Noel-Hirst, A., & Ford, C. (2024). Using incongruous genres to explore music making with AI generated content. In B. Bailey, C. Latulipe, & S. Ferguson (Eds.), *Proceedings of the 16th Conference on Creativity & Cognition* (pp. 229–240). Association for Computing Machinery. https://doi.org/10.1145/3635636.3656198

Chakrabarty, T., Laban, P., & Wu, C.-S. (2025). *Can AI writing be salvaged? Mitigating idiosyncrasies and improving human–AI alignment in the writing process through edits*. arXiv preprint arXiv:2409.14509

Chakrabarty, T., Padmakumar, V., Brahman, F., & Muresan, S. (2024). Creativity support in the age of large language models: An empirical study involving professional writers. In B. Bailey, C. Latulipe, & S. Ferguson (Eds.), *Proceedings of the 16th Conference on Creativity & Cognition* (pp. 132–155). Association for Computing Machinery. https://doi.org/10.1145/3635636.3656201

Chambers, R., Tack, N., Pearson, E., Martin, L. J., & Ferraro, F. (2024). BERALL: Towards generating retrieval-augmented state-based interactive fiction games. In P. Ammanabrolu, M.-A. Côté, B. Zou Li, L. J. Martin, N. V.

Peng, A. Suhr, R. Tamari, L. Teodorescu, A. Trischler, J. Weston, & X. Eric Yuan (Eds.), *The 4th Wordplay: When Language Meets Games Workshop* (pp. 1–11). https://2024.aclweb.org/program/workshops/

Chatman, S. B., & Chatman, S. (1980). *Story and discourse: Narrative structure in fiction and film*. Cornell University Press.

Chen, X., Aksitov, R., Alon, U., Ren, J., Xiao, K., Yin, P., Prakash, S., Sutton, C., Wang, X., & Zhou, D. (2023). *Universal self-consistency for large language model generation*. arXiv preprint arXiv:2311.17311

Chen, X., Jia, S., & Xiang, Y. (2020). A review: Knowledge reasoning over knowledge graph. *Expert Systems with Applications*, *141*(C), 112948. https://doi.org/10.1016/j.eswa.2019.112948

Czikszentmihalyi, M. (1990). *Flow: The psychology of optimal experience*. Harper & Row.

Edge, D., Trinh, H., Cheng, N., Bradley, J., Chao, A., Mody, A., Truitt, S., & Larson, J. (2024). *From local to global: A graph RAG approach to query-focused summarization*. arXiv preprint arXiv:2404.16130

Egri, L. (1972). *The art of dramatic writing: Its basis in the creative interpretation of human motives*. Simon and Schuster.

Englert, C. S., & Mariage, T. V. (1991). Shared understandings: Structuring the writing experience through dialogue. *Journal of Learning Disabilities*, *24*(6), 330–342. https://doi.org/10.1177/002221949102400602

Freytag, G. (1894). *Die Technik des Dramas*. S. Hirzel.

Gallotta, R., Todd, G., Zammit, M., Earle, S., Liapis, A., Togelius, J., & Yannakakis, G. N. (2024). Large language models and games: A survey and roadmap. *IEEE Transactions on Games*, *2024*, 1–18. arXiv preprint arXiv:2402.18659 https://doi.org/10.1109/TG.2024.3461510

Gao, W., Gu, X., Hu, B., Huang, H., Huang, M., Ke, P., Li, W., Liu, Y., Tang, J., Wang, H., Wen, B., Wu, L., Xu, J., & Zhou, J. (2024). Benchmarking complex instruction-following with multiple constraints composition. *Advances in Neural Information Processing Systems*, *37*, 137610–137645. https://doi.org/10.52202/079017-4371

Gao, Y., Xiong, Y., Gao, X., Jia, K., Pan, J., Bi, Y., Dai, Y., Sun, J., & Wang, H. (2025). *Retrieval-augmented generation for large language models: A survey*. arXiv preprint arXiv:2312.10997

Genette, G. (1980). *Narrative discourse: An essay in method*. Cornell UP.

Gingrich, B. (2021). *The pace of fiction: Narrative movement and the novel*. Oxford University Press.

Gómez-Rodríguez, C., & Williams, P. (2023). *A confederacy of models: A comprehensive evaluation of LLMs on creative writing*. arXiv preprint arXiv:2310.08433

Guan, J., Huang, F., Zhao, Z., Zhu, X., & Huang, M. (2020). A knowledge-enhanced pretraining model for commonsense story generation. *Transactions of the Association for Computational Linguistics*, *8*, 93–108. https://doi.org/10.1162/tacl_a_00302

Hall, S. (2019). Encoding—Decoding (1980). In C. Greer (Ed.), *Crime and media* (pp. 44–55). Routledge.

He, X., Tian, Y., Sun, Y., Chawla, N. V., Laurent, T., LeCun, Y., Bresson, X., Hooi, B., & Retriever, G. (2024). *Retrieval-augmented generation for textual graph understanding and question answering*. arXiv preprint arXiv:2402.07630

Hogan, A., Blomqvist, E., Cochez, M., D'amato, C., Melo, G. D., Gutierrez, C., Kirrane, S., Gayo, J. E. L., Navigli, R., Neumaier, S., Ngomo, A.-C. N., Polleres, A., Rashid, S. M., Rula, A., Schmelzeisen, L., Sequeda, J., Staab, S., & Zimmermann, A. (2022). Knowledge graphs. *ACM Computing Surveys*, *54*(4), 1–37. https://doi.org/10.1145/3447772

Holland, N. N. (1998). Reader-response criticism. *The International Journal of Psycho-Analysis*, *79*(Pt 6), 1203–1211.

Jiang, Y., Wang, Y., Zeng, X., Zhong, W., Li, L., Mi, F., Shang, L., Jiang, X., Liu, Q., & Wang, W. (2024). *FollowBench: A multi-level fine-grained constraints following benchmark for large language models*. arXiv preprint arXiv:2310.20410

Kim, Y., Chang, Y., Karpinska, M., Garimella, A., Manjunatha, V., Lo, K., Goyal, T., & Iyyer, M. (2024). *Fables: Evaluating faithfulness and content selection in book-length summarization*. arXiv preprint arXiv:2404.01261

Kommineni, V. K., König-Ries, B., & Samuel, S. (2024). *From human experts to machines: An LLM supported approach to ontology and knowledge graph construction*. arXiv preprint arXiv:2403.08345

Lewis, P., Perez, E., Piktus, A., Petroni, F., Karpukhin, V., Goyal, N., Küttler, H., Lewis, M., Yih, W-t., Rocktäschel, T., Riedel, S., & Kiela, D. (2020). Retrieval-augmented generation for knowledge-intensive NLP tasks. *Advances in Neural Information Processing Systems*, *33*, 9459–9474.

McKee, R. (1997). *Substance, structure, style, and the principles of screenwriting*. Alba Editorial.

Meyer, L.-P., Stadler, C., Frey, J., Radtke, N., Junghanns, K., Meissner, R., Dziwis, G., Bulert, K., & Martin, M. (2023). LLM-assisted knowledge graph engineering: Experiments with ChatGPT. In C. Zinke-Wehlmann & J. Friedrich (Eds.), *Working Conference on Artificial Intelligence Development for a Resilient and Sustainable Tomorrow* (pp. 103–115). Springer Fachmedien Wiesbaden.

Mizrahi, M., Kaplan, G., Malkin, D., Dror, R., Shahaf, D., & Stanovsky, G. (2024). State of what art? A call for multi-prompt LLM evaluation. *Transactions of the Association for Computational Linguistics*, *12*, 933–949. https://doi.org/10.1162/tacl_a_00681

Pan, Z., Benini, S., Grierson, M., Savardi, M., & Smith, T. J. (2025). Exploring the creative potential of AI in filmmaking. In S. Andolina, N. Bryan-Kinns, & S. F. Alaoui (Eds.), *Proceedings of the 2025 Conference on Creativity and Cognition* (pp. 24–30). Association for Computing Machinery.

Peng, B., Zhu, Y., Liu, Y., Bo, X., Shi, H., Hong, C., Zhang, Y., & Tang, S. (2024). *Graph retrieval-augmented generation: A survey*. arXiv preprint arXiv:2408.08921
Peng, X., Quaye, J., Rao, S., Xu, W., Botchway, P., Brockett, C., Jojic, N., DesGarennes, G., Lobb, K., Xu, M., Leandro, J., Jin, C., & Dolan, B. (2024). Player-driven emergence in LLM-driven game narrative. In *2024 IEEE Conference on Games (CoG)* (pp. 1–8). IEEE. https://2024.ieee-cog.org/about/#committee
Puxan-Oliva, M. (2024). Assessing narrative space: From setting to narrative environments. *Poetics Today*, *45*(1), 79–103. https://doi.org/10.1215/03335372-10938618
Queneau, R. (2013). *Exercises in style*. New Directions Publishing.
Rimmon-Kenan, S. (2003). *Narrative fiction: Contemporary poetics*. Routledge.
Ryan, M.-L. (2015). *Narrative as virtual reality 2: Revisiting immersion and interactivity in literature and electronic media*. JHU Press.
Shaer, O., Cooper, A., Mokryn, O., Kun, A. L., & Ben Shoshan, H. (2024). AI-augmented brainwriting: Investigating the use of LLMs in group ideation. In F. 'Floyd' Mueller, P. Kyburz, J. R. Williamson, C. Sas, M. L. Wilson, P. Toups Dugas, & I. Shklovski (Eds.), *Proceedings of the CHI Conference on Human Factors in Computing Systems* (pp. 1–17). Association for Computing Machinery.
Simon, N., & Muise, C. (2022). TattleTale: Storytelling with planning and large language models. In R. De Benedictis, S. Parkinson, M. Roveri, & S. Storandt (Eds.), *ICAPS Workshop on Scheduling and Planning Applications* (pp. 1–9). ICAPS.
Strachan, J. W. A., Albergo, D., Borghini, G., Pansardi, O., Scaliti, E., Gupta, S., Saxena, K., Rufo, A., Panzeri, S., Manzi, G., Graziano, M. S. A., & Becchio, C. (2024). Testing theory of mind in large language models and humans. *Nature Human Behaviour*, *8*(7), 1285–1295. https://doi.org/10.1038/s41562-024-01882-z
Subbiah, M., Zhang, S., Chilton, L. B., & McKeown, K. (2024). Reading subtext: Evaluating large language models on short story summarization with writers. *Transactions of the Association for Computational Linguistics*, *12*, 1290–1310. https://doi.org/10.1162/tacl_a_00702
Sun, Y., Li, Z., Fang, K., Lee, C. H., & Asadipour, A. (2023). Language as reality: A co-creative storytelling game experience in 1001 nights using generative AI. *Proceedings of the AAAI Conference on Artificial Intelligence and Interactive Digital Entertainment*, *19*(1), 425–434. https://doi.org/10.1609/aiide.v19i1.27539
Sun, Y., Tang, Y., Gao, Z., Pan, Z., Xu, C., Chen, Y., Qian, K., Wang, Z., Braud, T., Lee, C. H., & Asadipour, A. (2023). AI Nüshu: An exploration of language emergence in sisterhood through the lens of computational linguistics. In J. Kim & V. Szabo (Eds.), *SIGGRAPH Asia 2023 Art Papers* (pp. 1–7). Association for Computing Machinery.
Taveekitworachai, P., Abdullah, F., Gursesli, M. C., Dewantoro, M. F., Chen, S., Lanata, A., Guazzini, A., & Thawonmas, R. (2023). Breaking bad: Unraveling influences and risks of user inputs to ChatGPT for game story generation. In L. Holloway-Attaway & J. T. Murray (Eds.), *International Conference on Interactive Digital Storytelling* (pp. 285–296). Springer.
Todorov, T. (1971). The 2 principles of narrative. *Diacritics*, *1*(1), 37–44. https://doi.org/10.2307/464558
Touvron, H., Lavril, T., Izacard, G., Martinet, X., Lachaux, M.-A., Lacroix, T., Rozière, B., Goyal, N., Hambro, E., Azhar, F., Rodriguez, A., Joulin, A., Grave, E., & Lample, G. (2023). *Llama: Open and efficient foundation language models*. arXiv preprint arXiv:2302.13971
Touvron, H., Martin, L., Stone, K., Albert, P., Almahairi, A., Babaei, Y., Bashlykov, N., Batra, S., Bhargava, P., Bhosale, S., Bikel, D., Blecher, L., Ferrer, C. C., Chen, M., Cucurull, G., Esiobu, D., Fernandes, J., Fu, J., Fu, W., … Scialom, T. (2023). *Llama 2: Open foundation and fine-tuned chat models*. arXiv preprint arXiv:2307.09288
Venkatraman, S., Tripto, N. I., & Lee, D. (2024). *CollabStory: Multi-LLM collaborative story generation and authorship analysis*. arXiv preprint arXiv:2406.12665
Wang, B., Li, Y., Lv, Z., Xia, H., Xu, Y., & Sodhi, R. (2024). LAVE: LLM-powered agent assistance and language augmentation for video editing. In D. Wang & D. Bodoff (Eds.), *Proceedings of the 29th International Conference on Intelligent User Interfaces* (pp. 699–714). Association for Computing Machinery.
Wang, P. J., & Kreminski, M. (2024). *Guiding and diversifying LLM-based story generation via answer set programming*. arXiv preprint arXiv:2406.00554
Wang, Y., Lin, J., Yu, Z., Hu, W., & Karlsson, B. F. (2023). Open-world story generation with structured knowledge enhancement: A comprehensive survey. *Neurocomputing*, *559*(C), 126792. https://doi.org/10.1016/j.neucom.2023.126792
Wang, Y., Lipka, N., Rossi, R. A., Siu, A., Zhang, R., & Derr, T. (2024). Knowledge graph prompting for multi-document question answering. In M. Wooldridge, J. Dy, & S. Natarajan (Eds.), *Proceedings of the AAAI Conference on Artificial Intelligence* (pp. 19206–19214). AAAI Press. https://doi.org/10.1609/aaai.v38i17.29889
Wei, J., Wang, X., Schuurmans, D., Bosma, M., Xia, F., Chi, E., Le, Q. V., & Zhou, D. (2022). Chain-of-thought prompting elicits reasoning in large language models. *Advances in Neural Information Processing Systems*, *35*, 24824–24837. https://doi.org/10.5555/3600270.3602070
Woolson, R. F. (2005). Wilcoxon signed-rank test. *Encyclopedia of Biostatistics, 6*, 4739–4740. https://doi.org/10.1002/0470011815.b2a15177
Yang, D., & Jin, Q. (2024). *What makes a good story and how can we measure it? A comprehensive survey of story evaluation*. arXiv preprint arXiv:2408.14622

Yoo, T., & Cheong, Y.-G. (2024). Leveraging LLM-constructed graphs for effective goal-driven storytelling. In N. Zhang, T. Wu, M. Wang, G. Qi, H. Wang, H. Chen (Eds.), *CEUR Workshop Proceedings. CEUR-WS* (pp. 83–95). CEUR Workshop Proceedings.

Zamfirescu-Pereira, J. D., Wong, R. Y., Hartmann, B., & Yang, Q. (2023). Why Johnny can't prompt: How non-AI experts try (and fail) to design LLM prompts. In A. Schmidt, K. Väänänen, T. Goyal, P. O. Kristensson, A. Peters, S. Mueller, J. R. Williamson, & M. L. Wilson (Eds.), *Proceedings of the 2023 CHI Conference on Human Factors in Computing Systems* (pp. 1–21). Association for Computing Machinery.

Zhang, Y., Li, Y., Cui, L., Cai, D., Liu, L., Fu, T., Huang, X., Zhao, E., Zhang, Y., Chen, Y., Wang, L., Luu, A. T., Bi, W., Shi, F., & Shi, S. (2023). *Siren's song in the AI ocean: A survey on hallucination in large language models*. arXiv preprint arXiv:2309.01219

Zhao, Z., Song, S., Duah, B., Macbeth, J., Carter, S., Van, M. P., Bravo, N. S., Klenk, M., Sick, K., & Filipowicz, A. L. (2023). More human than human: LLM-generated narratives outperform human-LLM interleaved narratives. In B. Bailey & C. Latulipe (Eds.), *Proceedings of the 15th Conference on Creativity and Cognition* (pp. 368–370). Association for Computing Machinery.

Zhu, Y., Wang, X., Chen, J., Qiao, S., Ou, Y., Yao, Y., Deng, S., Chen, H., & Zhang, N. (2024). LLMs for knowledge graph construction and reasoning: Recent capabilities and future opportunities. *World Wide Web*, *27*(5), 58. https://doi.org/10.1007/s11280-024-01297-w

Zunshine, L. (2006). *Why we read fiction: Theory of mind and the novel*. Ohio State University Press.

## About the authors

**Zhijun Pan** (Aldrich) is a Research Fellow and PhD student at UAL Creative Computing Institute, researching AI-mediated storytelling and audience cognition. He was previously with The University of Edinburgh and Ubisoft, has published in leading peer-reviewed venues, and has contributed to Red Dot Design Award and Lumen Prize-winning projects.

**Antonios Andronis** is a Data Scientist and Machine Learning Lead specializing in cloud-based data platforms and applied AI across advertising, behavioral analytics and entertainment. He guides multidisciplinary teams and develops end-to-end machine-learning systems for complex real-world problems.

**Eva Hayek** is a creative developer at Charismatic.ai. She holds an MSc in Creative Computing from the University of the Arts London, where she explored generative AI for emergent gameplay. Her background spans quantitative development and creative technology for interactive storytelling.

**Oscar A. P. Wilkinson** is a Machine Learning Engineer with a background in Computer Science and Artificial Intelligence. He designs and implements machine learning-based systems, with strong experience in Python development and applied ML workflows.

**Ilya Lasy** is a PhD student at TU Wien and a Machine Learning Engineer at Charisma.ai. His research focuses on interpretability of large language models, neuro-symbolic AI and dialogue systems within creative and entertainment technologies.

**Annette Parry** is Director and COO of Charismatic.ai and a senior producer with over 20 years' experience delivering complex digital and technical projects. She specialises in coordinating multidisciplinary, globally distributed teams and managing technology-driven creative workflows.

**Guy Gadney** is CEO of Charismatic.ai and a digital innovation leader with senior roles across Penguin Books, the BBC, and other media organisations. He works at the intersection of storytelling and emerging technologies and serves on advisory boards including Sheffield DocFest and Innovate UK's Bridge AI group.

**Tim J. Smith** is Professor of Cognitive Data Science at the Creative Computing Institute (CCI), University of the Arts London, and head of the Cognition in Naturalistic Environments (CINE) Lab.

**Mick Grierson** is Professor and Research Leader in Creative Computing at UAL. His work spans machine learning, signal processing and creative AI, informing tools used by major media and technology organisations. He has led pioneering programmes and major research projects shaping the field of Creative Computing.